\documentclass[lettersize,journal]{IEEEtran}
\usepackage{lineno}
\usepackage{amsmath,amsfonts}
\usepackage{algorithmic}
\usepackage{algorithm}
\usepackage{array}
\usepackage{booktabs} 
\usepackage{multirow}
\usepackage[caption=false,font=normalsize,labelfont=sf,textfont=sf]{subfig}
\usepackage{textcomp}
\usepackage{stfloats}
\usepackage{url}
\usepackage{verbatim}
\usepackage{xcolor}
\usepackage{bbding}
\usepackage{graphicx}
\usepackage{cite}
\usepackage[most]{tcolorbox}
\begin{document}

\title{Reliable Reasoning Path:  \\ Distilling Effective Guidance for LLM \\ Reasoning with Knowledge Graphs}
\author{Yilin Xiao, Chuang Zhou, Qinggang Zhang, Bo Li, Qing Li,~\IEEEmembership{Fellow,~IEEE}, Xiao Huang
\thanks{Yilin Xiao, Chuang Zhou, Qinggang Zhang, Bo Li, Qing Li and Xiao Huang are with The Hong Kong Polytechnic University, Hong Kong SAR, China (e-mail: \{yilin.xiao, chuang-qqzj.zhou, qinggangg.zhang\}@connect.polyu.hk, comp-bo.li@polyu.edu.hk, csqli@comp.polyu.edu.hk, xiao.huang@polyu.edu.hk.).}
}

\markboth{Journal of \LaTeX\ Class Files,~Vol.~14, No.~8, August~2021}%
{Shell \MakeLowercase{\textit{et al.}}: A Sample Article Using IEEEtran.cls for IEEE Journals}


\maketitle

\begin{abstract}
Large language models (LLMs) often struggle with knowledge-intensive tasks due to a lack of background knowledge and a tendency to hallucinate. To address these limitations, integrating knowledge graphs (KGs) with LLMs has been intensively studied. Existing KG-enhanced LLMs focus on supplementary factual knowledge, but still struggle with solving complex questions. We argue that refining the relationships among facts and organizing them into a logically consistent reasoning path is equally important as factual knowledge itself. Despite their potential, extracting reliable reasoning paths from KGs poses the following challenges: the complexity of graph structures and the existence of multiple generated paths, making it difficult to distinguish between useful and redundant ones. To tackle these challenges, we propose the RRP framework to mine the knowledge graph, which combines the semantic strengths of LLMs with structural information obtained through relation embedding and bidirectional distribution learning. Additionally, we introduce a rethinking module that evaluates and refines reasoning paths according to their significance. Experimental results on two public datasets show that RRP achieves state-of-the-art performance compared to existing baseline methods. Moreover, RRP can be easily integrated into various LLMs to enhance their reasoning abilities in a plug-and-play manner. By generating high-quality reasoning paths tailored to specific questions, RRP distills effective guidance for LLM reasoning.
\end{abstract}


\section{Introduction}
\begin{figure}[t]
    \centering
    \includegraphics[width=\linewidth]{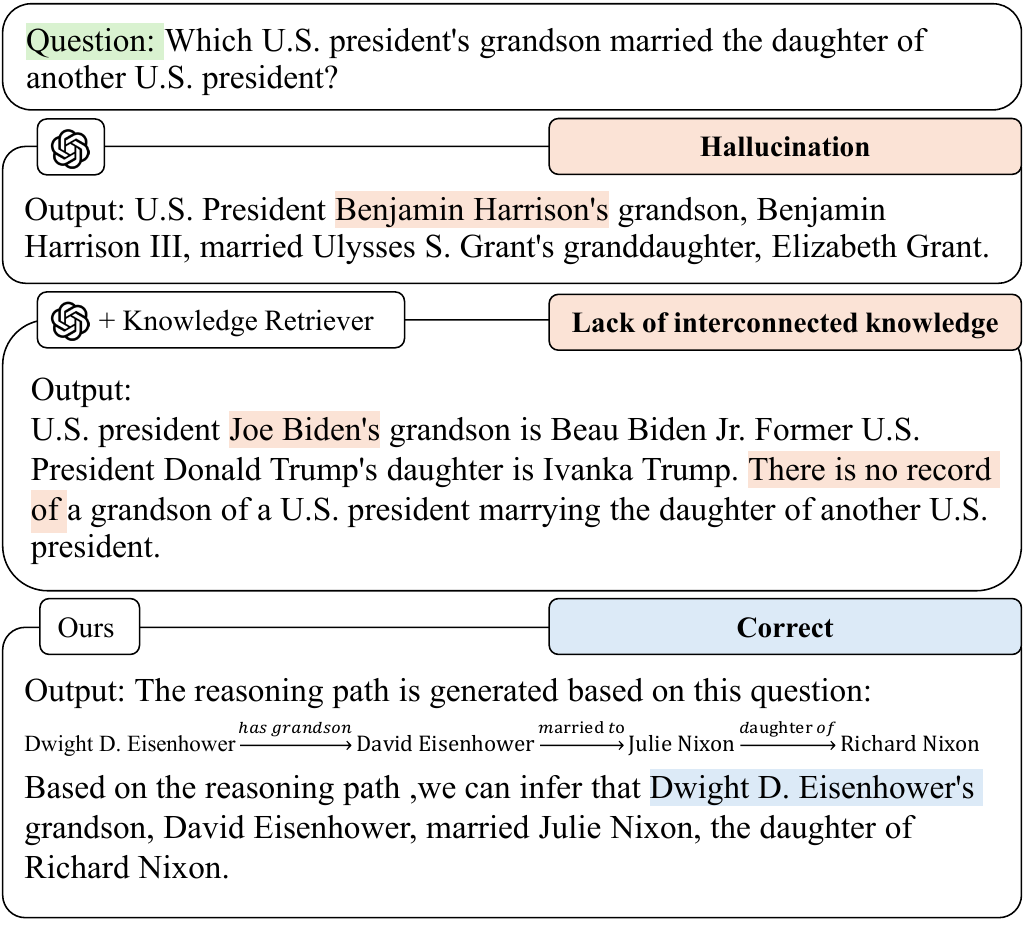}
    \caption{LLMs frequently hallucinate when addressing complex questions. Combining LLMs with a knowledge retriever accesses facts but struggles with questions requiring reasoning about interconnected knowledge. Our method bridges this gap by connecting knowledge through reasoning paths, enabling accurate solutions to complex questions.}
    \label{fig:0}
\end{figure}
\IEEEPARstart{L}{arge} language models (LLMs) have been widely used in real-world applications and demonstrated remarkable capabilities in various natural language understanding and generating tasks\cite{touvron2023llama,alpaca,cite_1,cite_2} due to their mighty semantic power derived from pre-training on the large corpus. However, LLMs still exhibit a tendency to hallucinate in knowledge-intensive tasks, particularly when they encounter questions that require external background knowledge or involve complex multi-hop reasoning~\cite{qinggang_survey,qinggang_knowgpt}. These hallucinations arise from the inherent property that they rely primarily on statistical correlations learned from large-scale corpora rather than a structured understanding of the underlying logical relationships. As a result, they may produce fluent yet factually incorrect or logically inconsistent responses. Recent efforts have focused on two main paradigms: fine-tuning LLMs with domain-specific data and employing retrieval-augmented generation (RAG) techniques that dynamically incorporate external knowledge during inference. However, fine-tuning may introduce new hallucinations by overfitting to spurious correlations within the data, rather than improving factual consistency in a generalizable way~\cite{korbak2022controlling,cite_3,cite_4,cite_5}. Furthermore, fine-tuning is computationally expensive and cannot be applied to real-time issues. Retrieval-augmented generation~\cite{cite_6,raptor,lightrag,graphrag} offers a promising alternative by supplementing LLMs with external knowledge at inference time. While this approach is effective for resolving relatively simple queries, such as providing definitions or clarifying terms, it often fails to deal with more complex reasoning tasks~\cite{santoro2016meta}. The retrieved documents are typically unstructured and disconnected, making it difficult for the model to construct a coherent and multi-step reasoning chain. To mitigate these issues, recent research has explored integrating LLMs with external knowledge sources, particularly knowledge graphs (KGs), which offer structured and factual representations of entities and their relationships~\cite{sun2024thinkongraph,luo2024reasoning,ijcai2024p734}. KG-enhanced LLMs have shown improved performance in factual recall and question answering by supplementing language models with relevant background knowledge based on entity connections.



However, while we acknowledge that such graphs can bring significant benefits for knowledge injection, we find that much of the existing work \cite{jiang2023unikgqa} focuses primarily on retrieving applicable content while overlooking the logic between knowledge. Beyond simple fact retrieval, the ability to synthesize a logically consistent and context-relevant reasoning path is essential for enabling robust LLM reasoning. For instance, answering a complex question may require combining multiple pieces of evidence through inferential steps that go beyond surface-level entity connections as shown in Figure \ref{fig:0}. KGs do not merely serve to provide knowledge units, they encapsulate the logic of reasoning as well. Mere retrieval of these independent knowledge units remains insufficient for addressing complex real-world problems, as such challenges demand integrative processing beyond isolated facts. Effective problem-solving in intricate scenarios hinges on the systematic organization of inter-unit relationships and the construction of reliable reasoning path, which together enable coherent knowledge traversal and logical inference.


Thus, we are motivated to organize relationships between knowledge units and facilitate the generation of reliable reasoning paths. But it's non-trivial to achieve due to two primary reasons. (1) Graph Complexity: These graphs' intricate connections and features often make it difficult to identify critical information. Despite the powerful semantic capabilities of LLMs, they cannot fully exploit complex structural information. For instance, questions that require a multi-hop reasoning path for accurate answers can be challenging for LLMs. (2) Accuracy of reasoning paths: we observe that existing methods generate numerous reasoning paths that are directly fed into LLMs. This approach is unreasonable because the correct reasoning paths may become obscured among the multitude, diminishing their effectiveness. Additionally, conflicting reasoning paths can introduce confusion within LLMs. Processing a large number of paths also reduces efficiency.

To tackle these challenges and distill effective guidance for LLMs, we introduce Reliable Reasoning Path (RRP), a novel framework that integrates LLMs and KGs to generate comprehensive and high-quality reasoning. RRP leverages relation embeddings and bidirectional distribution learning to extract structurally coherent and context-relevant reasoning paths. To further improve reasoning reliability, we introduce a rethinking module that evaluates and ranks the generated paths based on their contribution to answering the question. The resulting high-quality reasoning paths are then used to distill precise and interpretable guidance for LLMs, significantly enhancing their performance on reasoning tasks.

\textbf{Our main contributions are as follows:}
\begin{itemize}
\item  We propose a novel framework, \textbf{Reliable Reasoning Path (RRP)}, which tightly integrates large language models (LLMs) and knowledge graphs (KGs) to facilitate structured reasoning. RRP focuses not only on retrieving factual knowledge, but also on organizing it into coherent reasoning paths tailored to specific questions.

\item We develop a powerful reasoning path generation module that combines the semantic capabilities of LLMs with the structural priors of KGs. This module incorporates relation embedding to capture latent connections between entities and bidirectional distribution learning to ensure consistency and completeness of the generated paths, enabling more accurate multi-hop reasoning.

\item We introduce a rethinking modue that critically analyzes the generated reasoning paths from both structural and semantic perspectives, which selects the most informative ones based on relevance and logical coherence for LLM reasoning while eliminating redundant and invalid ones.

\item Experiments on two benchmark datasets show that our framework excels in reasoning task, outperforming all state-of-the-art methods while using LLM with only 7B parameters. We also assess the effect of each module of our model and the plug-and-play with various LLMs.
\end{itemize} 

\section{Related Work}

To customize LLMs for knowledge-intensive tasks, researchers have proposed various methods to enhance the reasoning ability of LLMs, falling into two main categories: Enhancing LLMs withot KG and KG-enhanced LLMs.

\subsection{Enhancing LLMs withot KG}
Early attempts focused on fine-tuning techniques to enhance LLM reasoning~\cite{yang2024self}. However, integrating new knowledge via fine-tuning can result in the model generating new hallucinations or even succumbing to catastrophic forgetting~\cite{korbak2022controlling}, especially when the new knowledge contradicts what was previously learned~\cite{gekhman2024does}. 

Recently, Retrieval-Augmented Generation models have been extensively explored to enhance LLMs with external knowledge from text corpora or online sources~\cite{lewis2020retrieval}. However, these approaches face challenges in knowledge-intensive tasks: $(i)$ These RAG documents may have varying levels of quality, accuracy, and completeness, leading to potential inconsistencies or errors in the retrieved knowledge~\cite{zhu2021retrieving}. $(ii)$ The lack of explicit relationships and structured organization limits the reasoning capabilities of RAG models, as they cannot leverage the structured connections to derive new insights or generate more contextually appropriate responses~\cite{santoro2016meta}.

\begin{figure*}[t]
    \centering
    \includegraphics[width=\linewidth]{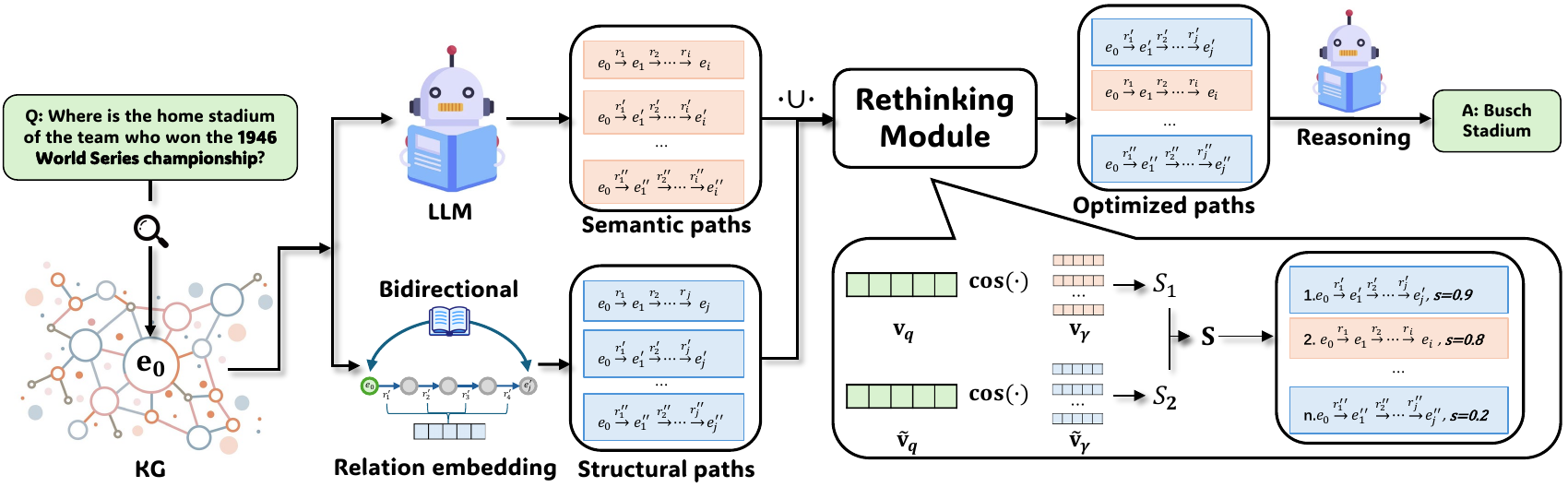}
    \caption{The overall framework of the proposed method. Given the question and KG, 1) we first utilize the powerful semantic capabilities of LLMs to generate reasoning paths that are semantically relevant for answering questions. 2) Second, employing the modules described in Section.\ref{sec.nsm}, we extract structural information from the KG to generate reasoning paths that are structurally related. 3) Finally, we combine these two sets of reasoning paths, prioritize them using the Rethinking module, and input the optimized paths to the LLMs for final reasoning.}
    \label{fig:main}
\end{figure*}

\subsection{KG-enhanced LLMs}

Earlier studies adopted a heuristic way to inject knowledge from KGs into the LLMs during pre-training or fine-tuning. ERNIE~\cite{sun2021ernie} incorporates entity embeddings and aligns them with word embeddings in the pre-training phase, encouraging the model to better understand and reason over entities. UniKGQA~\cite{jiang2023unikgqa} represents the first attempt to harness the capabilities of LLMs for jointly performing knowledge retrieval and multi-hop reasoning in a unified framework.


Another line of work focuses on retrieving relevant knowledge from KGs at inference time to augment the language model's context. Typically, K-BERT~\cite{liu2020k} uses an attention mechanism to select relevant triples from a KG based on the input context, which are then appended to the input sequence. More recently, KG prompting has been intensively studied for integrating factual knowledge into LLMs. KD-CoT~\cite{kddot} and KG-CoT~\cite{ijcai2024p734} build upon the concept of chain-of-thought, guiding LLMs through a step-by-step reasoning process while enabling timely correction of erroneous reasoning. Their factuality and faithfulness are validated using an external knowledge graph. RoG~\cite{luo2024reasoning} presents a planning-retrieval-reasoning framework that synergizes LLMs and KGs for more transparent and interpretable reasoning. SubgraphRAG~\cite{subgraphrag} integrates a lightweight MLP with a parallel triple-scoring mechanism for efficient and flexible subgraph retrieval. Despite their effectiveness, existing models focus on designing methods for knowledge retrieval or generation. They directly feed the retrieved or generated knowledge into the LLM for reasoning, which does not necessarily provide effective guidance to the LLMs. This is because LLMs often struggle to distinguish between valid and redundant knowledge.

\section{Preliminary}

\begin{itemize}
\item \textbf{Knowledge Graph (KG)}. KG represents structured knowledge in the form of a graph, where the information is encoded as a collection of triples: $\mathcal{G}=\left \{ \left \langle e,r,e^{'}|e,e^{'}\in \mathcal{E}, r\in \mathcal{R}  \right \rangle  \right \},$ where $\mathcal{E}$ and $\mathcal{R}$ denote the entity set and relation set, $e$ and $e^{'}$ are entities, and $r$ corresponds to a directed relation between the entities.
\item \textbf{Reasoning Paths}. Reasoning paths in KG can be viewed as instantiations of a relation path $\left \{ r_{1},r_{2},\dots,r_{n}   \right \}$, where a sequence of entities and relations forms a path. Specifically, a reasoning path can be represented as: $\gamma =e_{0}\overset{r_{1} }{\rightarrow} e_{1}\overset{r_{2} }{\rightarrow}\dots \overset{r_{n} }{\rightarrow}e_{n}$.

\item \textbf{LLM Reasoning with KGs} is a fundamental reasoning task that focuses on answering natural language questions by leveraging the structured knowledge stored in a KG. Given a natural language question $q$ and a KG $\mathcal{G}$, the objective of this task is to first accurately identify the entities referenced or implied in the question, and then generate the correct answer by reasoning over the relations and entities within the knowledge graph.
\end{itemize}

\section{Methodology}
In this section, we propose the RRP, which consists of three key components: 1) \emph{Semantic reasoning path generation}: This component leverages the powerful semantic capabilities of LLMs to generate reasoning paths related to the semantics of the given question in the KG. 2) \emph{Structural reasoning path generation}: To address the limitations of LLMs in capturing graph structural information, especially in multi-hop reasoning, where structural relationships in the graph may not be fully consistent with the semantics of the question. We leverage relation embedding and bidirectional distribution learning to mine the structural information of the knowledge graph. The reasoning paths generated by this component are structurally consistent with the question, thus complementing the semantic generation of LLMs. 3) \emph{Rethinking Module}:  This module optimizes the generated reasoning paths by eliminating redundancy, rearranging the paths, and prioritizing the most relevant ones. This step significantly distills effective guidance for LLM Reasoning.
\subsection{Semantic reasoning path generation} 
\label{sec.llm}
The entities in the KG are dynamically updated, which introduces uncertainty when the reasoning process is heavily dependent on a specific entity. In contrast, reasoning paths represent more stable and structured relations in KG. As a sequence of relations connecting entities, reasoning paths provide a solid foundation for reasoning tasks based on KGs. Therefore, we leverage the powerful reasoning capabilities of LLMs to generate reasoning paths. The goal of this module is to generate reasoning paths from KG that are as reliable as possible. To achieve this, we minimize the KL divergence between the posterior and the prior distributions of reliable reasoning paths. Given a question $q$ and an answer $a$, we identify $\gamma (q, a)$, a reasoning path on the KG that connects $e_{q}$ and $e_{a}$. This path can then be regarded as a reliable reasoning path for correctly answering $q$. The posterior distribution of reliable reasoning paths $\mathcal{P}(\gamma )$ is defined as follows:

\begin{equation}
\mathcal{P}(\gamma )\simeq \mathcal{P}(\gamma\mid a,q,\mathcal{G})=
\left\{
\begin{array}{@{}l@{}}
\frac{1}{\Gamma},\ \exists \gamma (e_{q},e_{a} )\in \mathcal{G}, \\
0,\ \mathrm{else}.
\end{array}
\right.
\label{eq1}
\end{equation}

where we assume a uniform prior distribution over the set of all reliable reasoning paths, denoted as $\Gamma$, and $\exists \gamma (e_{q},e_{a} )\in \mathcal{G}$ represent the existence of a reasoning path instance $\gamma$ in $\mathcal{G}$ connecting the question entity $e_{q}$ to the answer entity $e_{a}$. We define the probability of generating a reliable reasoning path $\gamma$ as $P_{\alpha } (\gamma \mid q)$, where $\alpha$ represents the parameters of LLMs. Based on these, the KL divergence can be computed as following formula:

\begin{equation}
\label{eq2}
\begin{aligned}
\mathcal{L}_{r_{\mathrm{1}}}
&= D_{\mathrm{KL}}(\mathcal{P}(\gamma\mid a,q,\mathcal{G})\left |  \right |P_{\alpha } (\gamma \mid q) ) \\
&= - \mathbb{E}_{\gamma \sim \mathcal{P}(\gamma\mid a,q,\mathcal{G})} \log P_\alpha(\gamma \mid q) + \text{CONST} \\
&= - \sum_{\gamma \in \Gamma^*} \mathcal{P}(\gamma\mid a,q,\mathcal{G}) \log P_\alpha(\gamma \mid q) + \text{CONST}, \\
&\simeq - \frac{1}{|\Gamma^*|} \sum_{\gamma \in \Gamma^*} \log P_\alpha(\gamma \mid q),
\end{aligned}
\end{equation}

where the exact computation of the expectations is intractable due to the large number of reasoning paths $\Gamma$, we approximate the expectations by considering the subset of shortest paths $\gamma \in \Gamma^*\subset \Gamma$ between $e_q$ and $e_a$ in KG, and CONST is omitted in the final optimization, as it does not contribute to the loss function. By optimizing Equation \ref{eq2}, we aim to maximize the likelihood of LLMs generating reliable reasoning paths. This process effectively distills knowledge from the KG into the LLMs.

\subsection{Structural reasoning path generation}
\label{sec.nsm}

Following the module described in Section~\ref{sec.llm}, we leverage the powerful semantic capabilities of LLMs to extract reasoning paths that are highly semantically relevant to the question. However, we argue that relying solely on semantic relevance is insufficient for accurate reasoning, particularly in multi-hop reasoning scenarios where structural relationships play a crucial role. Reasoning paths that are structurally related with the question can provide correct reasoning logic, enabling more robust reasoning. To address this limitation, we enhance the framework's structural mining capability by relation embedding and bidirectional distribution learning.

\begin{algorithm}[t]
    \caption{Structural reasoning path generation}
    \label{alg:algorithm_structural}
    \textbf{Input}: Question $q$, Knowledge Graph $\mathcal{G}$.\\
    \textbf{Output}: Reasoning Paths $\Gamma^*_{structural}$.\\
    \begin{algorithmic}[1] 
        \STATE $\Gamma^* \leftarrow \emptyset$.;
        \STATE  Employ GloVe to get initial embeddings $\mathbf{v}_q^{initial}$ of $q$.
        \STATE $\mathbf{v}_q^{initial}$ are processed through LSTM, where the final hidden state $\mathbf{v}_q$ represents the question.
        \STATE Initialize the entity embeddings $\mathbf{v}_e$ by: \\$\mathbf{v}_{e}^{0}  = \sigma\left( \sum_{\langle e, r, e^{'} \rangle \in \mathcal{G}} \mathbf{v}_{r} \cdot W_1 \right)$;
        \FOR{$i \gets 1$ \textbf{to} $n$}
        \STATE Construct a match vector $\mathbf{m}_{\langle e, r, e^{'} \rangle}^{i}$ by:
        \\$\mathbf{m}_{\langle e, r, e^{'} \rangle}^{i} = \sigma \left( \omega^{i} \odot W_2 \mathbf{v}_{r} \right)$;
        \STATE Aggregate matching messages from neighboring by:
        \\$\tilde{\mathbf{v}}_{e}^{i} = \sum_{\langle e, r, e^{'} \rangle \in \mathcal{G}} P_{e'}^{i-1} \cdot \mathbf{m}_{\langle e, r, e^{'} \rangle}^{i}$;
        \STATE Update the embeddings of entities as follows:
        \\$\mathbf{v}_{e}^{i} = \text{FFN}([\mathbf{v}_{e}^{i-1}; \tilde{\mathbf{v}}_{e}^{i}])$;
        \STATE Probability distribution $P^{i}$ could be obtained:
        \\$P^{i} = \text{softmax}\left((V^{i})^T \mathbf{W}\right)$;
        \STATE Update the loss $\mathcal{L}_{r_{\mathrm{2}}}$ based on $P^{i}$.
        \ENDFOR
        \STATE Obtain Reasoning Paths $\Gamma^*_{structural}$ based on $P^{n}$.
        \STATE \textbf{return} Reasoning Paths $\Gamma^*_{structural}$.
    \end{algorithmic}
\end{algorithm}

We begin by translating the input question into a sequence of reasoning instructions. Specifically, we employ GloVe \cite{glove} to derive the initial word embeddings of the question $q$. These embeddings are then processed through an LSTM encoder, where the final hidden state represents the question. Subsequently, a recurrent decoder is utilized to generate the corresponding reasoning instructions $\left \{ \omega^{i} \right \} _{i=1}^{n}$ after $n$ step. The $\left \{ \omega^{i} \right \} _{i=1}^{n}$ is then used as a guiding signal to facilitate the learning of the entity distribution. To initialize the entity embeddings $\mathbf{v}_e$, we incorporate information about the relations involving each entity $e$, defined as following formula:

\begin{equation}
    \mathbf{v}_{e}^{0}  = \sigma\left( \sum_{\langle e, r, e^{'} \rangle \in \mathcal{G}} \mathbf{v}_{r} \cdot W_1 \right),
\end{equation}
where $W_1$ represents the learnable weight matrix. In contrast to the conventional use of entity embeddings, we focus on the embedding of relations, denoted as $\mathbf{v}_{r}$. Our rationale is to effectively generate a more accurate and robust reasoning path, the emphasis should be placed on the relations between entities, as they capture the underlying structure and dependencies within the KG. Given a triple $\langle e, r, e^{'} \rangle$, we can construct a match vector $\mathbf{m}_{\langle e, r, e^{'} \rangle}^{i}$. This vector can be learned by matching the current reasoning instruction $\omega^{i}$ with the relation vector $\mathbf{v}_{r}$:

\begin{equation}
    \mathbf{m}_{\langle e, r, e^{'} \rangle}^{i} = \sigma \left( \omega^{i} \odot W_2 \mathbf{v}_{r} \right),
\end{equation}
where $W_2$ represents the learnable weight matrix. Next, we aggregate the matching messages from neighboring triples, assigning weights based on the attention they received in the previous reasoning step:
 
\begin{equation}
    \tilde{\mathbf{v}}_{e}^{i} = \sum_{\langle e, r, e^{'} \rangle \in \mathcal{G}} P_{e'}^{i-1} \cdot \mathbf{m}_{\langle e, r, e^{'} \rangle}^{i},
\end{equation}
where $P_{e'}^{i-1}$ denotes the probability assigned to entity $e^{'}$ in the previous step. This aggregation effectively captures the relational context of an entity within the KG, reinforcing its representation. Subsequently, we update the embeddings of entities as follows:

\begin{equation}
\mathbf{v}_{e}^{i} = \text{FFN}([\mathbf{v}_{e}^{i-1}; \tilde{\mathbf{v}}_{e}^{i}]), 
\label{upadted}
\end{equation}
where $\text{FFN}(\cdot )$ denotes a feed-forward neural network layer. This update mechanism effectively encodes the structure of the relation path into the entity embedding. The model maintains a probability distribution $P^{i}$ over candidate entities $P^{i}$. This distribution can be formally defined as follows:

\begin{equation}
   P^{i} = \text{softmax}\left((V^{i})^T \mathbf{W}\right),
\end{equation}
where $V^{i}$ is the embedding matrix at the $i$-step, and $\mathbf{W}$ represents the learnable weight matrix that derive the entity distribution $P^{i}$, and $V^{i}$ is updated by Eq. \ref{upadted}.

During training, we believe that bidirectional distribution learning plays a crucial role in mining accurate structural information and generating more comprehensive reasoning paths. To formalize this approach, we define the forward entity distribution from the $e_q$ to the $e_a$ as $P_{f}^{i}$, and the backward entity distribution from $e_a$ to $e_q$ as $P_{b}^{i}$. The core assumption underpinning this approach is that $P_{f}^{i}$ and $P_{b}^{i}$ should exhibit high similarity and consistency for a reasoning process to be stable and accurate. This assumption ensures that the bidirectional distributions mutually reinforce one another, enhancing the module's robustness. To implement this principle, we introduce a loss function to generate structured reasoning paths as follows:

\begin{align}
\mathcal{L}_{r_{\mathrm{2}}} &= D_{\mathrm{KL}} \left( P_f^{i} \, , \, P_f^{*} \right) + 
D_{\mathrm{KL}} \left( P_b^{i} \, , \, P_b^{*} \right)  \notag \\
&\quad + \sum_{i=1}^{n-1} D_{\mathrm{JS}}(P_f^{i}, P_b^{n-i}),
\end{align}    

where $D_{\mathrm{JS}}(\cdot )$ denotes the Jensen-Shannon divergence, a symmetric measure that quantifies the difference between two probability distributions. The pseudocode is provided in Algorithm~\ref{alg:algorithm_structural}.
\subsection{Rethinking the reasoning path}

In the generation of reasoning paths, our approach differs from prior work by ensuring that the reasoning paths not only cover semantically related paths but also cover structurally related paths. This comprehensive alignment enhances the overall quality and effectiveness of the generated reasoning paths. Previous methodologies often involve directly feeding all generated reasoning paths into LLMs to obtain the final answer. We argue that this practice is suboptimal, as merely including the correct reasoning path does not necessarily lead to the correct answer from the LLMs. This limitation arises because the correct reasoning path might be positioned at the end of the prompt or obscured by a multitude of redundant reasoning paths, which hinders the LLMs' ability to extract and utilize the pertinent knowledge effectively. Therefore, to further improve the efficacy of automated reasoning, we argue that a comprehensive global rethinking of the generated reasoning paths is crucial for improving reasoning accuracy.

\begin{algorithm}[t]
    \caption{RRP algorithm}
    \label{alg:algorithm}
    \textbf{Input}: Question $q$, Knowledge Graph $\mathcal{G}$.\\
    \textbf{Parameter}: $\lambda_{1}$, $\lambda_{2}$ and $\theta$.\\
    \textbf{Output}: Reasoning Paths $\Gamma^*$, Answer $a$.\\
    \begin{algorithmic}[1] 
        \STATE $\Gamma^* \leftarrow \emptyset$.;
        \STATE Gnerate semantic reasoning paths $\Gamma_{1}^{*}$;
        \STATE Gnerate structural reasoning paths $\Gamma_{2}^{*}$;
        \FOR{${\gamma}^{i} \in (\Gamma_{1}^{*}\cup \Gamma_{2}^{*})$}
        \STATE Compute the semantic similarity score $S_{1}(q, {\gamma}^{i})$;
        \STATE Compute the structural similarity score $S_{2}(q, {\gamma}^{i})$;
        \STATE Combine these socre by:\\
        $\mathbf{S}(q, {\gamma}^{i})=\lambda_{1}\cdot S_{1}(q, {\gamma}^{i})+\lambda_{2}\cdot S_{2}(q, {\gamma}^{i})$;
        \IF {$\mathbf{S}(q, {\gamma}^{i}) > \theta$}
        \STATE $\Gamma^*$.append$({\gamma}^{i})$;
        \ELSE
        \STATE Filter ${\gamma}^{i}$;
        \ENDIF
        \ENDFOR
        \STATE Sort$(\Gamma^*)$ by the $\mathbf{S}$;
        \STATE Feed $(q,\Gamma^*)$ to LLMs to get answer $a$;
        \STATE \textbf{return} answer $a$.
    \end{algorithmic}
\end{algorithm}

According to Zellig Harris's Distributional Hypothesis, linguistic units appearing in similar contexts tend to have similar meanings. In the context of LLMs, questions and their corresponding reasoning paths often appear within the same or similar contexts in the training corpus. Therefore, we assume that the embeddings of the question and its correct reasoning paths are similar in the feature space of the LLMs. Specifically, we utilize the module described in Section.\ref{sec.llm} to obtain the embedding $\mathbf{v}_q$ of the question and the embedding $\mathbf{v}_{\gamma}^{i}$ of the $i$-th reasoning path. Then the semantic similarity score between the question and the reasoning path is defined as following formula:
\begin{equation}
S_{1}(q, {\gamma}^{i}) = \cos\left( \mathbf{v}_q, \mathbf{v}_{\gamma}^{i} \right), 
\end{equation}
where \( \cos(\cdot) \) denotes the cosine similarity between two vectors. Furthermore, we assert that the structural similarity between the question and the reasoning path is crucial for generating accurate answers. As detailed in Section. \ref{sec.nsm}, our approach primarily leverages relational information when encoding questions. During the training process, we learn the valid reasoning paths between the question entity and the answer entity in a bidirectional manner. This bidirectional learning inherently embeds the structural information of the valid reasoning paths within the question representations. Consequently, we quantify the structural similarity score between the question and the reasoning path using the following formula:
\begin{equation}
S_{2}(q, {\gamma}^{i}) = \cos\left( \tilde{\mathbf{v}}_q, \frac{1}{n} \sum_{j=1}^{n} \tilde{\mathbf{v}}_{\gamma}^{{i}^{j}} \right), 
\end{equation}
where $\tilde{\mathbf{v}}$ donates the embedding obtained from the module in Section.\ref{sec.nsm}. Since the reasoning path comprises multiple entities, we intuitively represent the embedding of the path by averaging the embeddings of its constituent entities. Subsequently, we combine the semantic similarity score and the structural similarity score to quantify the overall importance of each reasoning path in accurately answering the question. The integration of these scores is computed as follows:
\begin{equation}
\mathbf{S}(q, {\gamma}^{i})=\lambda_{1}\cdot S_{1}(q, {\gamma}^{i})+\lambda_{2}\cdot S_{2}(q, {\gamma}^{i}),
\end{equation}
where $\lambda_{1}$ and $\lambda_{2}$ are the hyper-parameters. Utilizing the computed importance scores $\mathbf{S}(q, {\gamma}^{i})$, we sort the reasoning paths corresponding to the question $q$ in descending order of importance. We then set a threshold $\theta$ to eliminate reasoning paths that are not sufficiently relevant to $q$. Specifically, we eliminate paths whose importance scores fall below $\theta$. When preparing the input for the LLMs for question answering, we include the remaining reasoning paths in the prompt, arranged in order of decreasing significance. This approach ensures that LLMs focus on the most pertinent reasoning paths, enhancing the accuracy of the generated answers. For fine-tuning the LLMs, we follow the standard prompt tuning process \cite{luo2024reasoning}. To clearly and intuitively express the RRP algorithm, the pseudocode is provided in Algorithm~\ref{alg:algorithm}. The algorithm leverages the modules introduced in Section~\ref{sec.llm} and Section~\ref{sec.nsm} to generate comprehensive and accurate reasoning paths. These paths are then refined through a rethinking process, where they are reordered based on their significance, redundant paths are eliminated, and the optimized paths are subsequently fed into LLMs to enhance reasoning.

\section{Experiments}
\subsection{Datasets}
We evaluate the effectiveness of the proposed method on two widely used benchmark datasets. We follow the consistent settings as previous works~\cite{luo2024reasoning, jiang2023unikgqa} to use the same train and test split ratios for fair comparison.

\textbf{WebQuestionsSP (WebQSP)} \cite{webqsp} comprises 4737 natural language questions, each associated with the reliable reasoning paths within Freebase. This dataset focuses on questions that require straightforward reasoning paths.

\textbf{ComplexWebQuestions (CWQ)} \cite{cwq} consists of 34699 complex natural language questions. Unlike WebQSP, which primarily focuses on straightforward factual queries, CWQ emphasizes multi-hop reasoning, as the reliable reasoning paths often involve traversing multiple relationships in the KG.

\subsection{Implementation \& Evaluation}
For the module described in Section.\ref{sec.llm}, we utilize LLaMA2-Chat-7B as the backbone model. The training process is configured with the following hyperparameters: the number of epochs is set to 3, the batch size to is set 4, and the learning rate is set to $2e^{-5}$. For the module detailed in Section.\ref{sec.nsm}, the training is conducted with 80 epochs, a batch size of 40, and a learning rate of $4e^{-4}$. The hyperparameters for the Rethinking module are different for the two datasets: for WebQSP, the parameters $(\lambda_{1}, \lambda_{2})$ are set to $(0.5, 0.5)$, while for CWQ, they are set to $(0.1, 0.9)$. In line with previous studies, we adopt Hits@1 and F1 score as the primary evaluation metrics to assess the effectiveness of our method. Hits@1 evaluates the proportion of questions for which the top-1 predicted answer matches the ground truth. F1 score accounts for scenarios where a question corresponds to multiple correct answers. It considers a tradeoff between precision and recall rate, thereby providing a comprehensive measure of the model's coverage and accuracy.

\begin{table*}[!t]
\centering
\resizebox{0.80\textwidth}{!}{%
\begin{tabular}{llcccc}
\hline
\multirow{2}{*}{\textbf{Category}}         & \multirow{2}{*}{\textbf{Method}} & \multicolumn{2}{c}{\textbf{WebQSP}} & \multicolumn{2}{c}{\textbf{CWQ}} \\ \cline{3-6} 
                                  &                         & Hits@1        & F1         & Hits@1      & F1        \\ \hline
\multirow{5}{*}{Embedding-based}        & KV-Mem \cite{miller-etal-2016-key}                 & 46.7          & 34.5       & 18.4        & 15.7      \\
                                  & EmbedKGQA \cite{saxena-etal-2020-improving}               & 66.6          & -          & 45.9        & -         \\
                                  & NSM \cite{He_NSM}                     & 68.7          & 62.8       & 47.6        & 42.4      \\
                                  & TransferNet \cite{shi-etal-2021-transfernet}             & 71.4          & -          & 48.6        & -         \\
                                  & KGT5 \cite{saxena-etal-2022-sequence}                    & 56.1          & -          & 36.5        & -         \\ \hline
\multirow{4}{*}{Retrieval-based}        & GraftNet \cite{sun-etal-2018-open}               & 66.4          & 60.4       & 36.8        & 32.7      \\
                                  & PullNet \cite{sun-etal-2019-pullnet}                & 68.1          & -          & 45.9        & -         \\
                                  & SR+NSM \cite{zhang-etal-2022-subgraph}                  & 68.9          & 64.1       & 50.2        & 47.1      \\
                                  & SR+NSM+E2E \cite{zhang-etal-2022-subgraph}             & 69.5          & 64.1       & 49.3        & 46.3      \\ \hline
\multirow{5}{*}{LLMs}             & Flan-T5-xl \cite{JMLR:v25:23-0870}             & 31.0          & 19.9          & 14.7        & 13.2         \\
                                  & Alpaca-7B \cite{alpaca}              & 51.8          & 34.3          & 27.4        & 22.2         \\
                                  & LLaMA2-Chat-7B \cite{touvron2023llama}         & 64.4          & 28.0          & 34.6        & 16.9         \\
                                  & ChatGPT                 & 66.8          & 39.3          & 39.9        & 28.5         \\
                                  & ChatGPT+CoT \cite{cot}            & 75.6          & -          & 48.9        & -         \\ \hline
\multirow{7}{*}{LLMs+KGs}         & KD-CoT \cite{kddot}                 & 68.6          & 52.5       & 55.7        & -         \\
                                  & UniKGQA \cite{jiang2023unikgqa}                & 77.2          & 72.2       & 51.2        & 49.1      \\
                                  & ToG+ChatGPT \cite{sun2024thinkongraph}            & 76.2          & -          & 58.9        & -         \\
                                  & ToG+LLaMA2-70B  \cite{sun2024thinkongraph}        & 68.9          & -          & 57.6        & -         \\
                                  & KG-CoT \cite{ijcai2024p734}             & 84.9          & -       & 62.3        & -      \\
                                  & RoG \cite{luo2024reasoning}      & 85.7          & 70.8       & 62.6        & 56.2      \\ 
                                  & SubgraphRAG+ChatGPT \cite{subgraphrag}      & 83.1          & 69.2       & 56.3        & 49.1      \\ 
                                  & SubgraphRAG+LLaMA3-8B \cite{subgraphrag}      & 86.6          & 70.6       & 57.0        & 47.2      \\ 
                                  
                                  \cmidrule{2-6}
                                  & RRP(ours)                    & \textbf{90.0}          & \textbf{72.5}       & \textbf{64.5}        & \textbf{56.5}      \\ \hline
\end{tabular}%
}
\caption{Performance comparison with various baselines on the two standard datasets.}
\label{tab:main res}
\end{table*}

\subsection{Baselines}

We evaluate the proposed method by comparing it against strong baselines that represent a diverse range of categories.These baselines are categorized into four distinct types:

\textbf{Embedding-based Methods:} These methods focus on constructing high-quality vector representations of entities and relations to facilitate accurate reasoning and question-answering.
\begin{itemize}
\item KV-Mem \cite{miller-etal-2016-key} is a modified version of the memory network that enhances document readability by employing distinct encodings during the addressing and output stages of the memory read operation.
\item EmbedKGQA \cite{saxena-etal-2020-improving} introduces the use of KG embeddings for question answering, relaxing the requirement for answer selection to a pre-specified local neighborhood.
\item NSM \cite{He_NSM} employs a teacher-student framework that generates more reliable intermediate supervision signals, improving reasoning performance.
\item TransferNet \cite{shi-etal-2021-transfernet} enables multi-step reasoning by dynamically attending to different parts of the question at each step. It computes activation scores for relations and transfers entity scores along activated relations in a differentiable manner.
\item KGT5 \cite{saxena-etal-2022-sequence} demonstrates that an off-the-shelf encoder-decoder Transformer model can function as a scalable and versatile KGE model, achieving strong results in incomplete KG question answering tasks.
\end{itemize}

\textbf{Retrieval-based Methods:} Approaches that enhance the question-answering process by incorporating external retrieval mechanisms.

\begin{itemize}
\item GraftNet \cite{sun-etal-2018-open} introduces a directed propagation method inspired by personalized PageRank, commonly used in information retrieval.
\item PullNet \cite{sun-etal-2019-pullnet} employs an iterative process to construct a question-specific subgraph that captures information relevant to the question. During each iteration, a GCN identifies subgraph nodes to expand using retrieval operations on the corpus and/or knowledge base. Once the subgraph is constructed, a similar GCN extracts the final answer from the subgraph.
\item SR \cite{zhang-etal-2022-subgraph} is a trainable subgraph retriever that operates independently of the subsequent reasoning process. This decoupling enables a plug-and-play framework to enhance any subgraph-oriented model.
\end{itemize}

\textbf{Large Language Models (LLMs):} Approaches \cite{JMLR:v25:23-0870,alpaca,touvron2023llama,cot} that leverage pre-trained, large-scale neural language models to generate answers based on their extensive language understanding capabilities.

\textbf{LLMs+KGs Methods:} Hybrid approaches that integrate the interpretative capabilities of LLMs with the information from KGs to enhance reasoning accuracy.
\begin{itemize}
\item KD-CoT \cite{kddot} introduces a Knowledge-Driven Chain-of-Thought framework that verifies and modifies reasoning traces in CoT through interactions with external knowledge. This approach mitigates hallucinations and error propagation, improving reasoning accuracy.
\item UniKGQA \cite{jiang2023unikgqa} proposes a pre-trained language model-based approach that unifies retrieval and reasoning within both the model architecture and parameter learning, streamlining the question-answering process.
\item ToG \cite{sun2024thinkongraph} introduces a novel paradigm where an LLM agent iteratively executes beam search on a knowledge graph (KG), effectively discovering and exploring reasoning paths.
\item KG-CoT \cite{ijcai2024p734} builds on the Chain-of-Thought framework by incorporating a small-scale, step-by-step graph reasoning model to facilitate reasoning over knowledge graphs.
\item RoG \cite{luo2024reasoning} presents a planning-retrieval-reasoning framework. RoG generates relation paths, which are subsequently used to retrieve reasoning paths from knowledge graphs. These retrieved paths serve as input for LLMs to conduct more structured and accurate reasoning.
\item SubgraphRAG \cite{subgraphrag} introduces an innovative framework that combines a lightweight multilayer perceptron with a parallel triple-scoring mechanism for efficient and flexible subgraph retrieval, while encoding directional structural distances to enhance retrieval effectiveness. 
\end{itemize}

\subsection{Main results}
As presented in Table \ref{tab:main res}, we first analyze results by category. For traditional methods that rely on embeddings and methods that rely on retrieval, performance remains generally stable but exhibits clear bottlenecks. When LLMs are used alone for reasoning, their performance is not as strong as some embedding-based and retrieval-based methods. This is because, although LLMs have powerful semantic capabilities, they lack effective structural mining. However, methods that combine LLMs with KGs outperform other categories overall, as KGs provide structured knowledge that complements the deficiencies of LLMs in mining structural information. These results indicate that integrating LLMs with KGs constitutes an effective paradigm for reasoning tasks. Furthermore, our proposed method achieves state-of-the-art performance compared with other counterparts within the LLMs+KGs category, on both the WebQSP and CWQ datasets.

Specifically, on the WebQSP dataset, our method surpasses the recent RoG approach by 4.3 percentage points in Hits@1 and achieves a 1.7 percentage point improvement in the F1 score. This improvement can be attributed to the inclusion of more comprehensive reasoning paths in our method and the rethinking mechanisms of reasoning paths. In addition, our method surpasses methods from other categories by 15 to 60 points, further demonstrating the effectiveness of combining a large language model with a knowledge graph. Similarly, on the CWQ dataset, our method outperforms RoG by 2.0 percentage points in Hits@1, demonstrating its effectiveness in handling complex queries. Methods from other categories struggle on these challenging queries because such queries demand both high semantic understanding and deep structural understanding. Notably, these results were achieved using an LLM with only 7B parameters, demonstrating that our approach delivers competitive performance without relying on excessively large model sizes, making it a practical solution for resource-constrained scenarios.

\begin{table}[h!]
\centering
\resizebox{\columnwidth}{!}{%
\begin{tabular}{lcc|cc}
\toprule
\textbf{Methods} & \multicolumn{2}{c|}{\textbf{WebQSP}} & \multicolumn{2}{c}{\textbf{CWQ}} \\
                 & \textbf{Hits@1} & \textbf{Recall}   & \textbf{Hits@1} & \textbf{Recall} \\
\midrule
GPT-4o-mini                        & 67.08 & 40.64 & 41.43 & 35.95 \\
GPT-4o-mini + RRP         & \textbf{91.10} & \textbf{81.52} & \textbf{71.68} & \textbf{66.79} \\
\midrule
ChatGPT                        & 66.77 & 49.27 & 39.90 & 35.07 \\
ChatGPT + RRP         & \textbf{89.93} & \textbf{79.34} & \textbf{63.07} & \textbf{57.99} \\
\midrule
Alpaca-7B                      & 51.78 & 33.65 & 27.44 & 23.62 \\
Alpaca-7B + RRP       & \textbf{78.13} & \textbf{59.34} & \textbf{48.40} & \textbf{42.00} \\
\midrule
LLaMA2-Chat-7B                 & 64.37 & 44.61 & 34.60 & 29.91 \\
LLaMA2-Chat-7B + RRP  & \textbf{86.79} & \textbf{75.39} & \textbf{59.78} & \textbf{54.76} \\
\midrule
Flan-T5-xl                     & 30.95 & 17.08 & 14.69 & 12.25 \\
Flan-T5-xl + RRP      & \textbf{69.66} & \textbf{45.90} & \textbf{40.61} & \textbf{34.27} \\
\bottomrule
\end{tabular}}
\caption{Performance improvements of the proposed method across different LLMs.}
\label{tab:Plug-and-play}
\end{table}
\subsection{Plug-and-play study}

Because our method is compatible with any large language model during inference, we evaluate how much its integration improves reasoning performance across various models. To conduct this evaluation, we first use our framework to generate a set of high‐quality reasoning paths for each question. We then provide both the original question and the corresponding reasoning paths as input to different LLMs, allowing them to derive answers directly based on those paths without any additional fine tuning. Table \ref{tab:Plug-and-play} summarizes the results of this experiment. On the WebQSP dataset, integrating our method leads to substantial gains in Hits@1 for all tested models. For example, Flan-T5 xl achieves a 38.71 percent increase compared to its baseline performance without our reasoning paths. Other LLMs also experience improvements exceeding 20 percent in Hits@1. These results confirm that supplying structured reasoning chains significantly enhances the model’s ability to locate the correct answer even when the underlying language model remains unchanged. In the CWQ dataset, similar trends emerge: GPT-4o mini exhibits a 30.25 percent gain in Hits@1, and every other LLM tested shows an improvement of over 20 percent. This consistent enhancement across these datasets firmly demonstrates that our method effectively unlocks stronger inference capabilities under different model architectures.

It is particularly noteworthy that GPT-4o mini combined with our reasoning path generation framework delivers state of the art accuracy on both WebQSP and CWQ. This outcome highlights two important points. First, the reasoning paths produced by our approach provide critical context that large language models alone may not infer reliably from the query text. Second, because no fine tuning is required, our method offers a practical way to boost performance in resource‐constrained scenarios where retraining or adapting large models is infeasible. In summary, this plug-and-play study demonstrates that our reasoning path generation module functions as a versatile and effective enhancement across a wide range of large language models. It yields significant and consistent improvements in complex question answering tasks.

\subsection{Hyperparameter sensitivity analysis}
\begin{figure}[htbp]
  \centering
  \begin{minipage}{\linewidth}
    \centering
    \includegraphics[width=\linewidth]{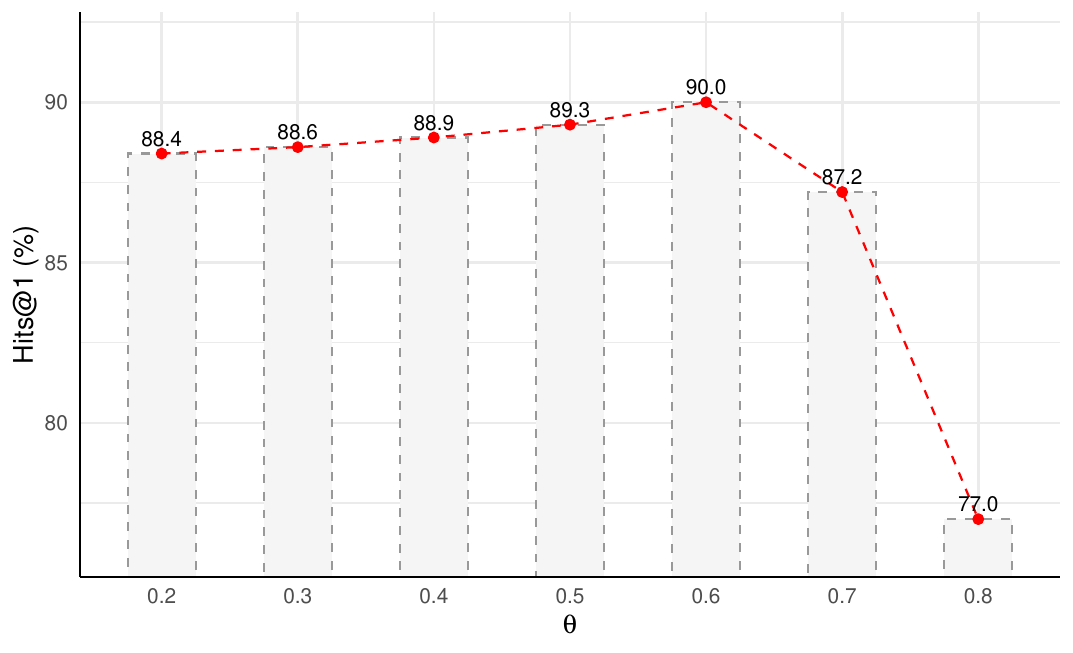}
    \caption{Performance across different values of $\theta$ on WebQSP.}
    \label{theta}
  \end{minipage}


  \begin{minipage}{\linewidth}
    \centering
    \includegraphics[width=\linewidth]{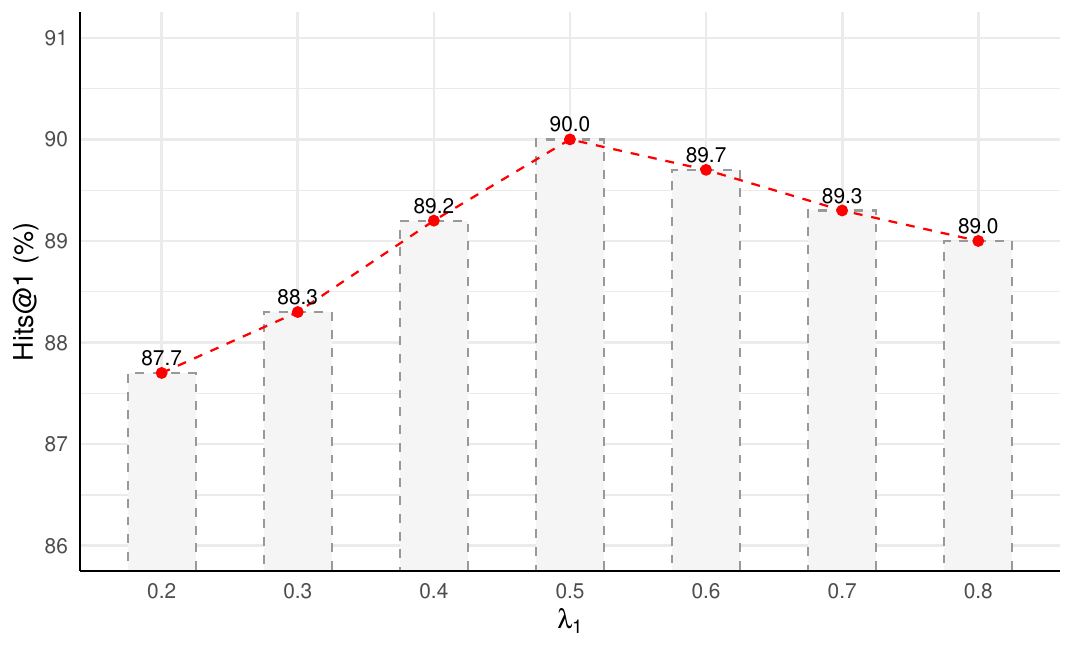}
    \caption{Performance across different values of $\lambda_1$ on WebQSP.}
    \label{lambda}
  \end{minipage}

\end{figure}
The proposed model includes two hyperparameters, $\theta$ and $\lambda$. The parameter $\theta$ denotes the threshold for filtering redundant reasoning paths; a lower threshold retains a greater number of paths. The parameter $\lambda$ controls the balance between semantic information and structural information. Specifically, $\lambda_{1}$ represents the weight of semantic information, $\lambda_{2}$ = 1 – $\lambda_{1}$ represents the weight of structural information, and together they satisfy $\lambda_{1}$ + $\lambda_{2}$ = 1. To thoroughly evaluate the sensitivity of our method to various hyperparameters, we conducted a comprehensive series of experiments.

Figure~\ref{theta} illustrates the effect of varying $\theta$. As $\theta$ increases from its minimum up to 0.6, model performance gradually improves and reaches its optimum at $\theta$ = 0.6. Beyond this point, further increases in $\theta$ cause performance to decline, with a marked drop observed when $\theta$ = 0.8. This behavior can be explained as follows: increasing $\theta$ more aggressively filters out unrelated reasoning paths, thereby reducing their negative impact on correct reasoning. At $\theta$ = 0.6, the model achieves an optimal balance, effectively removing the vast majority of irrelevant paths while retaining most of the valid ones. However, if the filtering threshold becomes too high, a large proportion of reasoning pathsare discarded, which includes some valid ones, leading to degraded performance.

Figure~\ref{lambda} shows the results for varying $\lambda_{1}$ (and consequently $\lambda_{2}$). As $\lambda_{1}$ increases from its minimum up to 0.5, model performance again improves and reaches its peak at $\lambda_{1}$ = 0.5. Further increases in $\lambda_{1}$ result in decreased performance. This trend arises because, for the WebQSP dataset, semantic information and structural information are equally important. As $\lambda_{1}$ rises, the model assigns more weight to semantics, which initially enhances performance; at $\lambda_{1}$ = 0.5, the weights for semantic and structural information are perfectly balanced. Beyond $\lambda_{1}$ = 0.5, structural information is weighted too lightly, which negatively affects the model’s ability to leverage structural cues and consequently reduces overall performance.

\subsection{Robustness to different sizes of KG}
\begin{table}[h]
\resizebox{\columnwidth}{!}{%
\begin{tabular}{lcc}
\hline
\multirow{2}{*}{\textbf{Characteristics}}& \multicolumn{2}{c}{\textbf{Knowledge Graph}} \\ \cline{2-3} 
                       & Freebase subgraph    & Wiki-Movie   \\ \hline
Number of entities     & 2,566,291            & 43,234       \\
Number of relations    & 7,058                & 9            \\
Number of triples      & 8,309,195            & 133,582      \\
Accuracy               & Medium               & High         \\
Average inference time & 2.9s                 & 2.6s         \\ \hline
\end{tabular}}
\caption{Comparison of knowledge graphs used to validate model robustness.}
\label{Comparison of KGs}
\end{table}

In the main results, we first demonstrate the effectiveness of the proposed method on WebQSP and CWQ, both of which are built upon the large-scale Freebase knowledge graph. However, since the sizes of knowledge graphs in real‐world applications can vary substantially, we further evaluate the robustness of our approach by incorporating additional knowledge graphs of different scales for comparison.

As shown in Table \ref{Comparison of KGs}, Freebase contains 2,566,291 entities, 7,058 relations, and a total of 8,309,195 triples, serving as a representative large‐scale knowledge graph. Because it spans multiple domains and includes a vast amount of information, its overall data quality is assessed as medium. In summary, Freebase is both large‐scale and highly complex. In contrast, Wiki‐Movie comprises 43,234 entities, 9 relations, and 133,582 triples, representing a medium‐scale knowledge graph. Since Wiki‐Movie is confined to a single domain and was constructed with high precision, its data quality is classified as high. Overall, Wiki‐Movie is of moderate scale and exhibits lower structural complexity compared to Freebase.

\begin{table}[htbp]
\resizebox{\columnwidth}{!}{%
\begin{tabular}{lcc}
\hline
\multirow{2}{*}{\textbf{Methods}} & \textbf{WebQSP(Freebase)} & \textbf{MetaQA-3(Wiki-Movie)} \\ \cline{2-3} 
                         & \multicolumn{2}{c}{Hits@1}                                  \\ \hline
RRP(ours)                & \textbf{90.0}                       & \textbf{99.5}                           \\
RoG                      & 85.7                       & 89.0                           \\
GraftNet                 & 66.4                       & 77.7                           \\
PullNet                  & 68.1                       & 91.4                           \\
EmbedKGQA                & 66.6                       & 94.8                           \\ \hline
\end{tabular}}
\caption{Robustness of proposed method to different KGs.}
\label{Robustness of RRP to different KG}
\end{table}

For comparative experiments, we selected a set of representative baseline methods; the results are reported in Table \ref{Robustness of RRP to different KG}, where WebQSP corresponds to the Freebase evaluation and MetaQA‐3 corresponds to the Wiki‐Movie evaluation. Across both large‐scale and medium‐scale knowledge graphs, our proposed method achieves the highest performance by a substantial margin over all baselines. This result further validates the robustness of RRP in realistic settings. Additionally, we computed the average inference time of our method on both knowledge graphs and found that the difference between them is negligible. This observation further confirms that the RRP framework consistently maintains robust computational efficiency, demonstrating scalability and stable performance regardless of knowledge graph scale.

\subsection{Ablation study}

\begin{table}[h]
\centering
\resizebox{\columnwidth}{!}{%
\begin{tabular}{cccccc}
\toprule
\textbf{Semantic} & \textbf{Structural} & \textbf{Rethinking} & \textbf{Hits@1} & \textbf{F1} \\
\midrule
\Checkmark & \Checkmark & \Checkmark & \textbf{90.0} & \textbf{72.5} \\
\Checkmark & \Checkmark & \XSolid          & 88.3 & 71.2 \\
\Checkmark & \XSolid          & \Checkmark & 86.7 & 69.5 \\
\XSolid          & \Checkmark & \Checkmark & 77.4 & 67.9 \\
\Checkmark & \XSolid          & \XSolid          & 84.8 & 68.6 \\
\XSolid          & \Checkmark & \XSolid          & 74.2 & 67.3 \\
\bottomrule
\end{tabular}%
}
\caption{Ablation study of the proposed method on WebQSP.}
\label{tab:ablation}
\end{table}

As presented in Table \ref{tab:ablation}, we conducted an ablation study to assess the contribution of each component in our proposed framework. The results indicate that the Semantic Reasoning Path Generation module yields strong performance on its own, owing to the advanced reasoning capabilities of LLMs. In contrast, the Structural Reasoning Path Generation module is designed to explore multi hop paths by traversing the explicit connections in the knowledge graph. By following graph edges, it uncovers chains of relations that link the query entities to candidate answers. However, this focus on structural exploration can cause it to overlook simple and direct reasoning paths that involve only one or two hops. In other words, it may miss short paths that are semantically rich but not obvious from purely structural heuristics.

When these two modules are combined, their strengths complement each other. The semantic component brings in flexibility and broader coverage, while the structural component ensures that important multi hop connections are considered. The integrated model therefore has access to a more comprehensive and accurate set of reasoning paths, encompassing both direct reasoning and longer, multi-step inferences. Moreover, adding the Rethinking module further improves performance when used alongside either or both of the previous components. This module performs a second pass over the candidate reasoning paths, scoring and ranking them to identify the most relevant chains. By assigning higher weights to paths that contribute positively to the final answer and reducing the influence of redundant or less relevant paths, the Rethinking module mitigates noise and enhances overall reasoning accuracy.

\subsection{Case study}

\begin{table*}[t]
\centering
\resizebox{1.0\textwidth}{!}{%
\begin{tabular}{p{4cm}|p{13cm}}
\hline
\textbf{Question} & The newspaper Zerkalo Nedeli is circulated in an area that has what as the official language? \\
\hline
\textbf{Reasoning Paths of RoG} & 
Zerkalo Nedeli $\to$ \textit{book.periodical.language} $\to$ \textbf{English Language} \\ 
& Zerkalo Nedeli $\to$ \textit{book.periodical.language} $\to$ \textbf{Russian Language} \\
& Zerkalo Nedeli $\to$ \textit{book.periodical.language} $\to$ \textbf{Ukrainian Language} \\
\hline
\textbf{Reasoning Paths of RRP} & 
\textbf{Reasoning paths are arranged in descending order of significance:}\\&
\textbf{Path 1(structural):} Zerkalo Nedeli $\to$ \textit{periodicals.newspaper\_circulation\_area.newspapers} $\to$ Ukraine $\to$ \textit{location.country.languages\_spoken} $\to$ \textbf{Ukrainian Language} \\ 
& \textbf{Path 2 (semantic):} Zerkalo Nedeli $\to$ \textit{book.periodical.language} $\to$ \textbf{Ukrainian Language} \\
& \textbf{Path 3 (semantic):} Zerkalo Nedeli $\to$ \textit{book.periodical.language} $\to$ \textbf{Russian Language} \\
& \textbf{Path 4 (semantic):} Zerkalo Nedeli $\to$ \textit{book.periodical.language} $\to$ \textbf{English Language} \\
\hline
\textbf{Outputs} & 
\textbf{RoG:} Based on the reasoning paths provided, the official language of the area where Zerkalo Nedeli is circulated is \textcolor{red}{Russian Language.} \\ 
& \textbf{RRP:} Based on the reasoning paths provided in order of significance, the official language of the area where Zerkalo Nedeli is circulated is \textcolor{blue}{Ukrainian Language.} \\
\hline
\textbf{Ground Truth} & \textbf{Ukrainian Language.} \\
\hline
\end{tabular}}
\caption{Case study for RRP. RRP can generate more comprehensive reasoning paths and rethink them by significance, allowing LLMs to obtain more effective guidance.}
\label{tab:case}
\end{table*}

As shown in Table \ref{tab:case}, this case study instantiates the superiority of RRP. Firstly, for this question, the SOTA method only generates the "periodical language" of "Zerkalo Nedeli". In contrast, the RRP successfully generates the language of the area where the newspaper is circulated, which is more relevant to the context of the question. This indicates that LLMs alone may be insufficient in accurately understanding the complex knowledge requirements and the logical interdependencies inherent in such questions. RRP effectively uncovers the logical chains that connect relevant knowledge, generating more comprehensive and accurate reasoning paths. Secondly, based on the reasoning paths, the SOTA method provides an incorrect answer, whereas the RRP accurately addresses the question by prioritizing the paths according to their significance. In general, the comprehensive and logically ordered reasoning paths generated by the RRP framework serve as distilled and targeted guidance for LLM reasoning.

\subsection{Expert-designed prompt template}
The semantic reasoning path generation module aims to generate valid reasoning path which can be guidance for answering the question. The prompt template is presented as shown in Figure ~\ref{fig:appendix_1}, where \textless Question\textgreater indicates the content of the question. 

\begin{figure}[h]
\begin{tcolorbox}[colback=gray!10, colframe=black, title= Semantic Paths Generation Prompt]
Please generate the valid reasoning paths that can be helpful for answering the following question:

\vspace{5pt}

\textless Question\textgreater
\end{tcolorbox}
\caption{Prompt template for semantic path generation.}
\label{fig:appendix_1}
\end{figure}

The LLM reasoning process serves as the final step after we get a comprehensive and well-ordered reasoning path using the RRP framework. By inputting the reasoning paths and the question, the LLM is guided to generate the correct answer. For more details, the prompt template is presented as shown in Figure ~\ref{fig:appendix_2}, where \textless Reasoning Paths \textgreater indicate the well-ordered instances of reasoning paths.

\begin{figure}[h]
\begin{tcolorbox}[colback=gray!10, colframe=black, title= LLM Reasoning Prompt]
Instructions: 

Please use the reasoning paths provided below to answer the question. The reasoning paths are listed in order of importance, with the first being the most important. Your task is to derive the simplest possible answer and return all potential answers as a list.

\vspace{5pt}

Reasoning Paths:

\textless Reasoning Paths\textgreater

\vspace{5pt}

Question:

\textless Question\textgreater
\end{tcolorbox}
\caption{Prompt template for large language model reasoning.}
\label{fig:appendix_2}
\end{figure}

\section{Conclusion}

In this paper, we present RRP, a novel approach to leveraging knowledge graphs for enhanced reasoning with large language models. Our framework strategically combines the semantic strengths of LLMs with rich structural information derived through relation embedding and bidirectional distribution learning, ensuring both the breadth and precision of the reasoning process. To further bolster reasoning quality, we introduce a rethinking module that systematically organizes, evaluates, and filters candidate reasoning paths, enabling the identification of the most coherent and impactful path for downstream LLM inference. Extensive experiments on two public datasets demonstrate that RRP consistently achieves state-of-the-art performance, surpassing a wide range of competitive baselines. Importantly, we showcase that these gains are realized even when employing relatively lightweight LLMs with only 7 billion parameters, underscoring the efficiency and scalability of our approach. Beyond empirical effectiveness, RRP’s modular, plug-and-play design provides seamless compatibility with any existing LLM architecture, facilitating easy integration into diverse real-world applications without necessitating extensive retraining or model modifications. Looking ahead, RRP lays a solid foundation for future advancements in knowledge-grounded language understanding. Its capacity to distill high-quality reasoning paths tailored to specific queries not only mitigates hallucination and enhances factual correctness, but also opens avenues for more transparent and explainable AI. We sincerely anticipate that the core ideas underpinning our framework will bring insights and inspire subsequent research on sophisticated knowledge-graph augmentation techniques and more robust reasoning paradigms in the evolving landscape of large-scale language intelligence.

\bibliographystyle{IEEEtran}
\bibliography{ref.bib}

\begin{IEEEbiography}[{\includegraphics[width=1in,height=1.25in,clip,keepaspectratio]{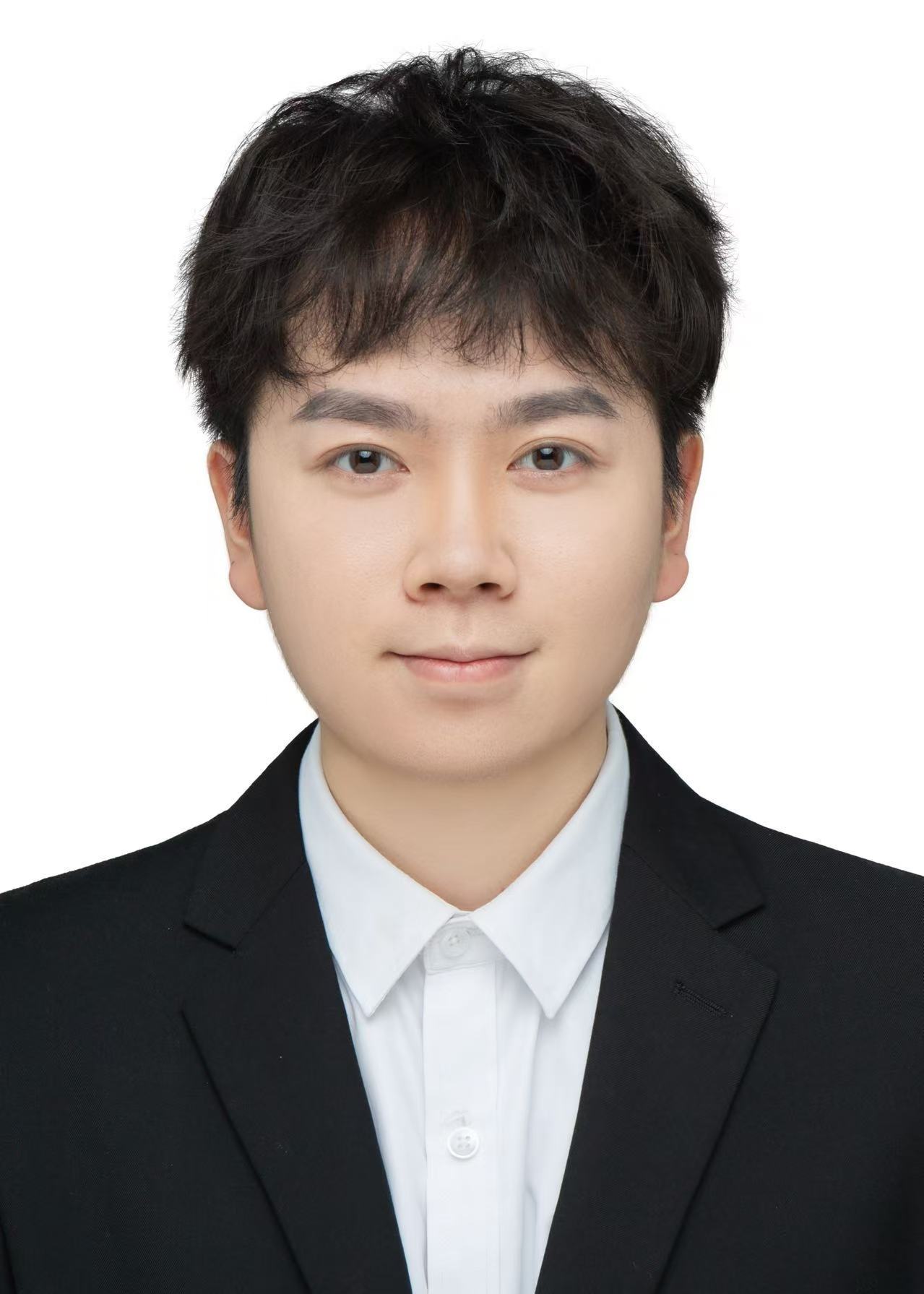}}]{Yilin Xiao} is a first-year Ph.D. candidate at the Department of Computing, The Hong Kong Polytechnic University. Before that, he received a bachelor's degree in Dalian University of Technology and a master's degree in Wuhan University. His research interests include KGs, reasoning, LLMs, and RAG. His works have been published in venues such as ACL, ACM Multimedia, IEEE TIV, IEEE TGRS and etc. He also serves as a reviewer for ACM Multimedia, IPM and etc.
\end{IEEEbiography}

\begin{IEEEbiography}[{\includegraphics[width=1in,height=1.25in,clip,keepaspectratio]{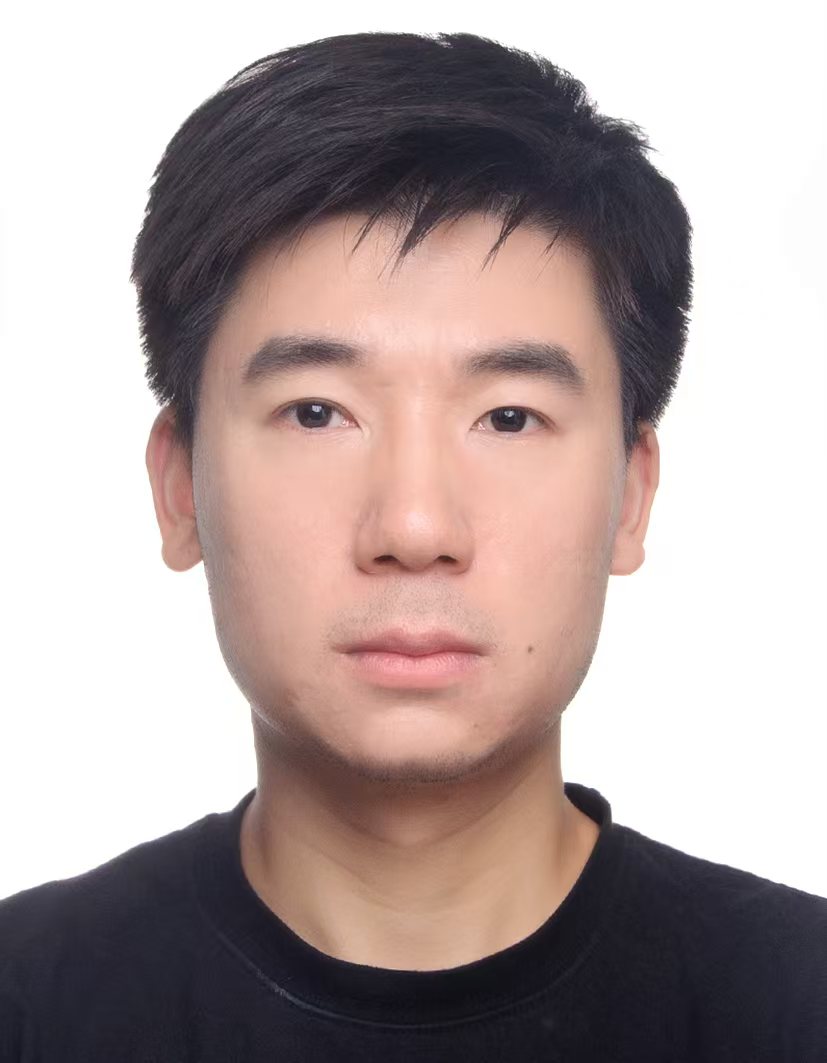}}]{Chuang Zhou} is a third-year Ph.D. candidate at the Department of Computing, The Hong Kong Polytechnic University.   Before that, he received a bachelor's degree in Economics and Statistics from  University College London and a master's degree in Statistics from The London School of Economics and Political Science. He is currently a member of the DEEP Lab supervised by Dr. Xiao Huang. His research interests include LLMs, recommendation, and GraphRAG. He has published several papers, including ACL, EMNLP, COLING and etc.
\end{IEEEbiography}

\begin{IEEEbiography}[{\includegraphics[width=1in,height=1.25in,clip,keepaspectratio]{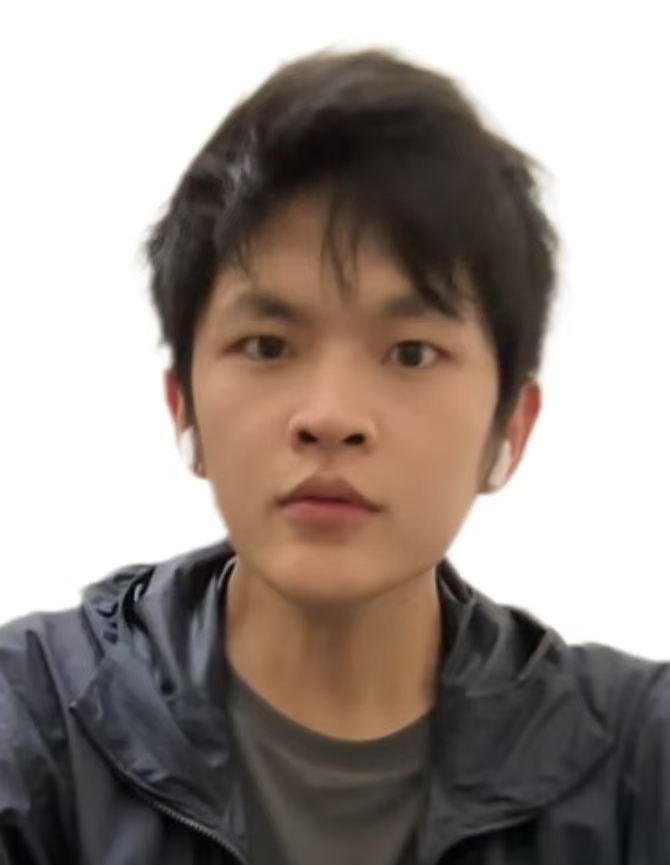}}]{Qinggang Zhang} is a fourth-year Ph.D. candidate at the Department of Computing, The Hong Kong Polytechnic University.   Before that, he received a bachelor's degree in Engineering in Computer Science from  Northwestern Polytechnical University. He is currently a member in DEEP Lab supervised by Dr. Xiao Huang. His research interests include KGs, LLMs, RAG and Text-toSQL. He has published over 20 papers while serving as reviewers for NeurIPS, ICML, ICLR,  KDD, IEEE TKDE and IEEE TPAMI.
\end{IEEEbiography}

\begin{IEEEbiography}[{\includegraphics[width=1in,height=1.25in,clip,keepaspectratio]{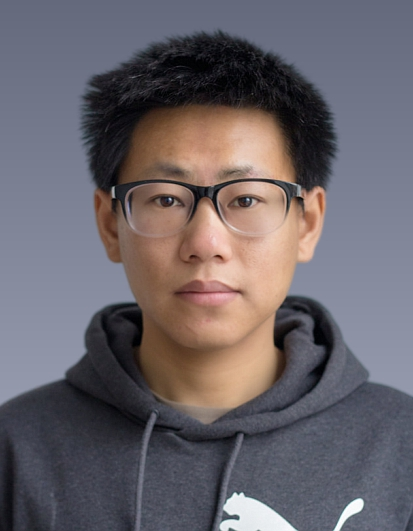}}]{Bo Li} is an assistant professor in the Department of Computing at the Hong Kong Polytechnic University. He received his Ph.D. in Computer Science from Stony Brook University. He is broadly interested in algorithms, AI, and computational economics, including problems related to resource allocation, game theory, online algorithms, and their applications to Blockchain and machine learning.
\end{IEEEbiography}

\begin{IEEEbiography}[{\includegraphics[width=1in,height=1.25in,clip,keepaspectratio]{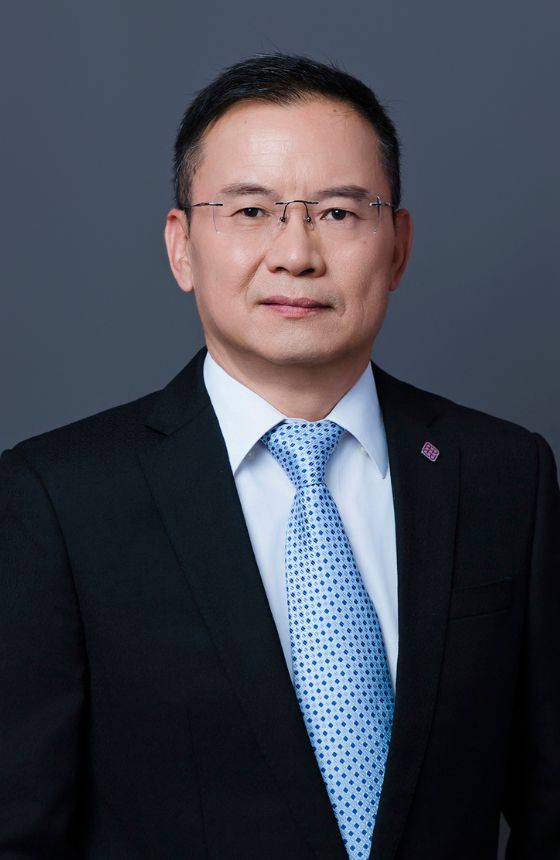}}]{Qing Li}(Fellow, IEEE) received the BEng degree from Hunan University, Changsha, China, and the MSc and PhD degrees from the University of Southern California, Los Angeles, all in computer science. He is currently a chair professor (Data science) and the head of the Department of Computing, Hong Kong Polytechnic University. He is a fellow of IET, a member of ACM SIGMOD and IEEE Technical Committee on Data Engineering. His research interests include object modeling, multimedia databases, social media, and recommender systems. He has been actively involved in the research community by serving as an associate editor and reviewer for technical journals, and as an organizer/co-organizer of numerous international conferences. He is the chairperson of the Hong Kong Web Society, and also served/is serving as an executive committee (EXCO) member of IEEE-Hong Kong Computer Chapter and ACM Hong Kong Chapter. In addition, he serves as a councilor of the Database Society of Chinese Computer Federation (CCF), a member of the Big Data Expert Committee of CCF, and is a Steering Committee member of DASFAA, ER, ICWL, UMEDIA, and WISE Society.
\end{IEEEbiography}

\begin{IEEEbiography}[{\includegraphics[width=1in,height=1.25in,clip,keepaspectratio]{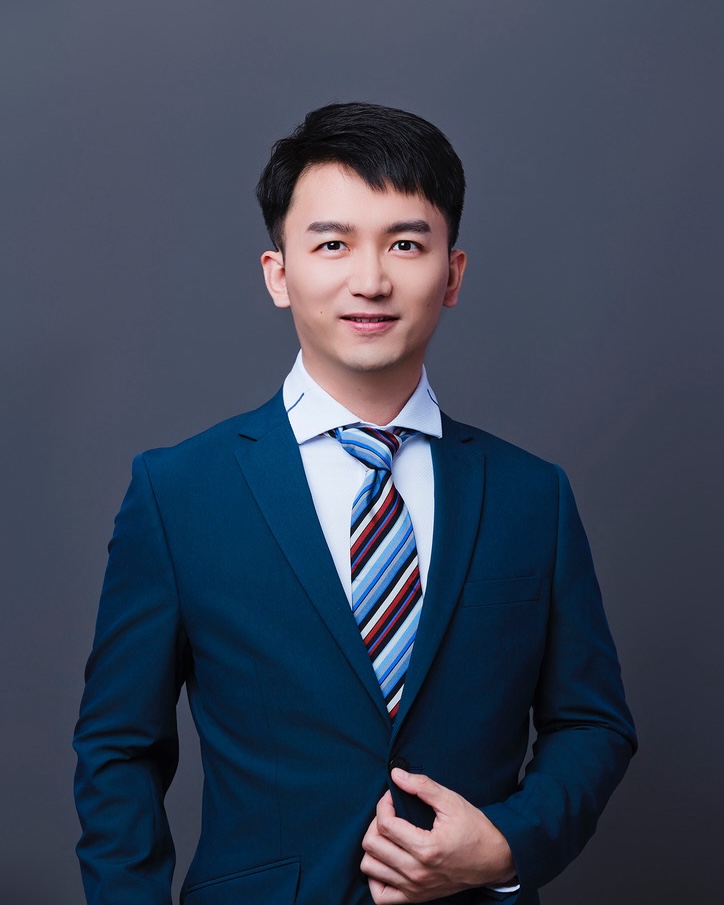}}]{Xiao Huang} is an Assistant Professor in the Department of Computing at The Hong Kong Polytechnic University. He earned his Ph.D. in Computer Engineering from Texas A\&M University in 2020, an M.S. in Electrical Engineering from the Illinois Institute of Technology in 2015, and a B.S. in Engineering from Shanghai Jiao Tong University in 2012. His scholarly contributions are highly regarded within the academic community, amassing over 4,900 citations. He received the Best Paper Award Honorable Mention at SIGIR 2023. He has led or completed seven research projects as Principal Investigator, securing funding exceeding 5 million HKD. He serves as a PhD Symposium Chair for ICDE 2025.
\end{IEEEbiography}

\end{document}